\documentclass{article}
\usepackage{ijcai26}

\usepackage{times}
\usepackage{soul}
\usepackage{url}
\usepackage[hidelinks]{hyperref}
\usepackage[utf8]{inputenc}
\usepackage[small]{caption}
\usepackage{graphicx}
\usepackage{amsmath}
\usepackage{amsthm}
\usepackage{booktabs}
\usepackage[switch]{lineno}

\usepackage{enumitem}

\newtheorem{definition}{Definition}

\usepackage{multirow}
\usepackage{xspace}
\usepackage[linesnumbered,ruled,vlined]{algorithm2e}
\usepackage{xcolor}

\newcommand{\tool}{\textsc{Alpha\-MapleSAT}\xspace}
\newcommand{\march}{\textsc{march}\xspace}
\newcommand{\March}{\textsc{March}\xspace}
\newcommand{\MC}{\textsc{Math\-Check}\xspace}
\newcommand{\minspeedupx}{1.61$\times$\xspace}
\newcommand{\maxspeedupx}{7.57$\times$\xspace}

\newcommand{\smsspeedupx}{7.57$\times$\xspace}

\DeclareMathOperator{\prop}{prop}

\title{AlphaMapleSAT: An MCTS-based Cube-and-Conquer SAT Solver for Hard Combinatorial Problems}

\author{
Piyush Jha$^1$\thanks{Equal contributions}\and
Zhengyu Li$^1$\footnotemark[1]\and
Zhengyang Lu$^2$\and
Raymond Zeng$^1$\and\\
Curtis Bright$^3$\And
Vijay Ganesh$^1$
\affiliations
$^1$Georgia Institute of Technology, USA\\
$^2$University of Waterloo, Canada\\
$^3$University of Windsor, Canada
\footnotetext[1]{Equal contributions}
}

\begin{document}

\maketitle

\begin{abstract}

This paper introduces \tool, a Cube-and-Conquer (CnC) parallel SAT solver that integrates Monte Carlo Tree Search (MCTS) with deductive feedback to efficiently solve challenging combinatorial SAT problems. Traditional lookahead cubing methods, used by solvers such as \march, limit their search depth to reduce overhead often resulting in suboptimal partitions. By contrast, \tool performs a deeper MCTS search guided by deductive rewards from SAT solvers. This approach enables informed exploration of the cubing space while keeping cubing costs low.

We demonstrate the efficacy of our technique via extensive evaluations against the widely used and established \march cubing solver on three well-known challenging combinatorial benchmarks, including the minimum Kochen--Specker (KS) problem from quantum mechanics, the Murty--Simon Conjecture, and the Ramsey problems from extremal graph theory. We compare \tool against \march using different types of conquering solvers such as SAT Modulo Symmetries (SMS) and SAT+CAS, both built on top of the \textsc{CaDiCaL} SAT solver. We show that in all cases, there is a speedup in elapsed real time (wall clock time) ranging from \minspeedupx to \maxspeedupx on a 128 core machine for the above-mentioned problems. We also perform cube-level and parallel scaling analysis over 32, 64, and 128 cores, which shows that \tool outperforms \march on all these settings. Our results show that deductively-guided MCTS search technique for cubing in CnC solvers can significantly outperform \march on hard combinatorial problems.
\end{abstract}

\section{Introduction}

In recent years, we have witnessed many hard combinatorial problems such as the Boolean Pythagorean Triples~\cite{heule2016solving}, Schur number five~\cite{heule2018schur}, and Lam's problem~\cite{bright2021sat} solved by SAT solving techniques. Among these, the Cube-and-Conquer (CnC) SAT solving approach has emerged as a dominant parallelization strategy~\cite{heule2011cube}. The method utilizes two distinct kinds of solvers: a lookahead ``cubing'' or ``splitting'' solver responsible for partitioning the input Boolean formula into sub-formulas using cubes (a conjunction of literals), and a subsequent ``conquering'' or ``worker'' solver (e.g., a conflict-driven clause learning SAT solver~\cite{silva1996grasp}) tasked with solving each sub-formula. CnC techniques~\cite{biere2021look} have demonstrated superior performance compared to sequential, portfolio, or traditional divide-and-conquer solvers~\cite{nejati2020cdcl} when dealing with hard combinatorial instances obtained from diverse applications in geometry~\cite{bright2021sat}, physics~\cite{likochenspecker}, and cryptography~\cite{zaikin2022inverting}. 

Further, previous work has demonstrated that sequential solvers alone are insufficient for solving these combinatorially hard problems. For example, taking the minimum Kochen--Specker (KS) problem as a representative, sequential runtime increases rapidly as the problem order $n$ grows. Sequential SAT+CAS solvers require over 32 hours to solve order 21~\cite{likochenspecker} and sequential SMS solvers take over 26 hours for the same instance~\cite{kirchweger2023co}, and both time out for $n=22$ after 5 days. Hence, one has no choice but to turn to using parallel solvers for such hard combinatorial problems. 

Furthermore, leading portfolio solvers like \textsc{Painless}~\cite{le2017painless} also fail on large instances and higher orders of hard combinatorial problems, even with the SAT+CAS solver integrated as a conquering solver. For example, they run out of memory on order 22 of the Kochen--Specker problem since the instance is too large.

It is well-known that the success of CnC solvers depends crucially on the order in which the variables are split. Consequently, lookahead cubing solvers are designed to use empirically tested metrics (e.g., propagation rate) to rank variables in an input formula and split variables in the order given by such a ranking resulting in sub-formulas that are expected to be easy to solve with conquering solvers. Note that a cubing solver may produce many cubes (each containing several literals) given an input formula. It does this typically by splitting on multiple variables and propagating them in a single run, resulting in such cubes.

Despite the dramatic success of CnC solvers, significant challenges remain. For example, the cubes created by traditional cubing solvers, such as \march~\cite{heule2011cube}, are not optimal, where optimality for cubes can be defined in terms of minimizing the total elapsed real time of the solver.\footnote{We define total elapsed real time as the time taken by a (parallel) CnC solver to solve the input formula and produce a SAT/UNSAT result.
See Section~\ref{metrics} for a more formal definition.}
Finding better cubes seems to require a deeper lookahead search, which can get prohibitively expensive as the size of the input formula increases. We define the problem we address in this paper as follows:

\vspace{0.1cm}
\noindent{\bf Problem Statement.} Design and implement a lookahead cubing solver that takes as input a Boolean formula $F(x_1,x_2,\dotsc,x_n)$ in conjunctive normal form (CNF), outputs a set $C=\{C_1,\dotsc,C_k\}$ of cubes\footnote{A cube is a conjunction of literals.} defined over the variables $\{x_1,\dotsc,x_n\}$, where the sub-formulas $F \wedge C_i$ are solved by conquering solvers (possibly in parallel) such that the total CPU and elapsed real time for cubing and solving $F$ is minimized.

\vspace{0.1cm}
\noindent{\bf \tool: An MCTS-based Cubing Solver.} To address the above problem, we present \tool, a novel cubing solver based on MCTS~\cite{coulom2006efficient} with deductive rewards, resulting in a heuristic search technique that is both cost-effective and enables an informed exploration of the cubes. In recent years, the popularity of MCTS has grown significantly, especially because of its success in solving hard combinatorial search problems for two-player games such as Go, Chess, and Shogi~\cite{silver2017mastering}. This has prompted numerous endeavors to apply MCTS to tackle combinatorial optimization problems from several domains including mathematics. 

\tool leverages the power of MCTS to peek ahead in the search space of splitting trees associated with the input formula, prioritizing splits that, while not necessarily offering the immediate highest reward, hold the potential to unlock better ones at subsequent splitting depths. \tool makes use of a deductive reward signal using an automated reasoning tool, a departure from traditional reward signals/functions which are typically provided by human domain experts in the context of MCTS or reinforcement learning. Deductive rewards, derived from automated reasoning tools (e.g., solvers), are very general and can be used to construct richer reward functions by augmenting human understanding of a problem domain. This informed exploration, guided by deductive reward signals, steers the search process towards promising partitions, significantly reducing the total elapsed time spent on both cubing and solving. 

The use of deduction-based metrics in the context of lookahead solvers is not new. In fact, \march uses SAT solvers to compute propagation rate and other deductively-derived metrics to rank variables to split on. What is new here is the combination of deductive reward with MCTS during selection and rollout, resulting in a powerful splitting heuristic.

\vspace{0.1cm}
\noindent{\bf Scope.} We evaluate on a small set of structured combinatorial benchmarks that have been used in prior CnC and \march-based studies and previously shown to outperform sequential and portfolio solvers. These benchmarks allow us to assess the effect of improved cubing decisions within the CnC paradigm, which is the focus of this work, rather than to target general-purpose or industrial SAT solving.

\subsection*{Contributions}
\begin{enumerate}[leftmargin=1.2em]
    \item We introduce \tool,\footnote{Cubing tool: \url{https://github.com/piyush-J/AlphaMapleSAT}\\ \hspace*{1.85em}Parallel CnC pipeline: \url{https://github.com/BrianLi009/AlphaMapleSAT-CnC}} a novel cubing solver integrating MCTS to overcome the limitations of both greedy (limited search used in current cubing solvers such as \march, resulting in sub-optimal cubes) and exhaustive search approaches for cubing (search cost overwhelms any benefits derived from optimal cubes). MCTS strikes a balance between immediate rewards and the potential for future rewards, leading to significantly faster cubing and solving times while keeping the cost of searching for good cubes low.
    
    \item \tool departs from traditional reward signals/functions in the context of MCTS, in that it uses an automated reasoning tool (a solver) to compute a deductive feedback reward signal (propagation rate). This novel approach is applied during the selection and rollout phases of MCTS\@. 
    
    \item We demonstrate the effectiveness of \tool by comparing it with the state-of-the-art \march cubing solver on benchmark problems such as the minimum Kochen--Specker~\cite{likochenspecker,kochen1990problem}, the Murty--Simon Conjecture~\cite{satmsym,caccetta1979diameter}, and the Ramsey problem~\cite{liramsey,ramsey1987problem}. We replace \march with \tool in the cubing phase and use different conquering solvers such as SAT+CAS and SMS, which in turn are built on top of \textsc{CaDiCaL}~\cite{Biere2024}. We show that in all cases, there is a speedup in elapsed real time from \minspeedupx to \maxspeedupx on a single node with 128 cores, highlighting {\tool}'s potential as a significant improvement over the existing state-of-the-art. We also perform parallel scaling analysis on 32, 64 and 128 cores, and cube-level analysis, explaining how improved cube quality translates into lower solving time in large-scale parallel settings.
    
\end{enumerate}

\section{Related Work}

\subsection{Cube-and-Conquer (CnC) Solvers}
The Cube-and-Conquer (CnC) SAT solving paradigm~\cite{heule2011cube}, combines a lookahead cubing phase with a conquering phase to address challenging combinatorial problems that are often intractable for traditional CDCL solvers. CnC is a subclass of divide-and-conquer approaches~\cite{nejati2020cdcl} that leverages lookahead heuristics to generate informed partitions of the original problem. \March remains the most widely adopted CnC cubing tool for over 15 years~\cite{heule2018schur,bright2022satisfiability,semenov2023using,likochenspecker}, and is integrated into modern solvers~\cite{heisinger2022paracooba,manthey2023parallel}, selecting split variables based on deductive metrics and propagation counts.

\subsection{MCTS}
The utilization of MCTS gained momentum following its triumph in two-player games~\cite{silver2017mastering}. MCTS has been applied in the context of SAT solving by introducing a UCT-based method for branching in the DPLL algorithm~\cite{previti2011monte}, with follow-up work improving reward functions via learned clauses and activity scores~\cite{schloeter2017monte,keszocze2020improving}.

More recently, Monte Carlo Forest Search~\cite{cameron2022monte} was introduced that combines MCTS with neural networks to identify candidate search trees for DPLL branching policies through an offline learning approach to avoid the computational cost of the reward function.
MCTS has also been applied to finding backdoors in MILP problems~\cite{khalil2022finding}, synthesizing SMT strategies~\cite{z3alpha}, Constraint Programming solvers~\cite{loth2013bandit}, and symbolic regression~\cite{symphylearner_2023}. 

Alongside this, several works integrate MCTS with neural network models for combinatorial optimization and satisfiability problems. For instance, CombOpt Zero~\cite{abe2019solving} leverages graph neural networks (GNNs) within the AlphaZero framework to solve NP-hard graph problems. GNN-based MCTS has also been explored for tackling the Quantified Boolean Formula Satisfaction (QSAT) problem~\cite{xu2022towards}, and similarly, MCTS combined with neural guidance has been explored in the context of SMT~\cite{graham2018guiding}. For a broader survey, we refer our readers to a recent review paper~\cite{swiechowski2023monte}, which provides a detailed overview of MCTS and its applications.

In contrast to these approaches, \tool frames the SAT cubing problem as a tree MDP but deliberately avoids neural network guidance. Instead, it uses a deductive reward signal computed via a solver during the selection and rollout phases of MCTS\@. This enables solver-guided, lightweight, yet informed search. To the best of our knowledge, \tool is the first cubing tool to integrate deductive reasoning within MCTS in an online setting without any learning or offline training.

\section{Background}

\subsection{Monte Carlo Tree Search (MCTS)}

MCTS is a heuristic search algorithm designed for navigating complex combinatorial spaces, particularly those represented as search trees~\cite{coulom2006efficient}. It has proven highly effective in reinforcement learning-based systems such as AlphaGo and AlphaZero~\cite{silver2017mastering}, which achieved breakthroughs in games like Go, Chess, and Shogi. In such two-player games, MCTS incrementally builds an $n$-ary search tree where the root node represents the current game state, edges correspond to valid actions, and child nodes represent resulting states. Terminal states are associated with scalar reward values.

The algorithm operates by iteratively performing the following four-step simulation process:
\textbf{(1) Selection:} Starting from the root, a tree policy selects actions that balance exploitation and exploration, typically using the Upper Confidence bound for Trees (UCT) formula~\cite{kocsis2006bandit}, until a leaf node is reached.
\textbf{(2) Expansion:} If the leaf node is non-terminal and has unvisited children, one or more child nodes corresponding to unexplored actions are added to the tree.
\textbf{(3) Rollout:} A simulation is executed from the new node to a terminal state using a default (often random or heuristic) policy.
\textbf{(4) Backup:} The result of the simulation is back-propagated through the visited nodes to update statistics such as average reward and visit count.

This process repeats until a specified computational budget (e.g., time or iteration limit) is exhausted. At that point, the action with the highest expected reward or most visits is selected. In two-player games, this process continues as the opponent responds and the game state updates. For additional details, we refer readers to comprehensive surveys on MCTS and its applications~\cite{browne2012survey,swiechowski2023monte}.

\subsection{Cube-and-Conquer (CnC) Solvers}
\label{bg_cnc}

The CnC SAT solving approach was introduced as a means to address challenging combinatorial problems that are often intractable for traditional CDCL solvers~\cite{heule2011cube}. CnC solvers follow a two-step process: a {\it lookahead cubing solver} first partitions the input formula into smaller subproblems, and a {\it conquering or worker solver} (e.g., a CDCL SAT solver) then solve these subproblems independently. The goal is to minimize the total elapsed time required to partition and solve the input SAT instance.

\begin{definition}[\bf Cube]
A cube is a conjunction of literals, e.g., $x_1 \wedge \dotsb \wedge x_n$, where each $x_i$ is a literal from the input formula.
\end{definition}

\begin{definition}[\bf Propagation rate] \label{def:prop_rate}
Given a formula\/ $F$ and a cube $c$, the propagation rate of\/ $F \wedge c$ is defined as the ratio between the number of propagations performed by a Boolean Constraint Propagation (BCP)~\cite{davis1962machine} on $F \wedge c$ and the size of\/ $c$. This metric serves as a widely used and effective heuristic for measuring the influence of variable assignments during cubing.
\end{definition}

\begin{definition}[\bf Lookahead heuristics]
Lookahead heuristics iterate over all variables (or candidate cubes) in a given formula, simplify the formula with respect to the given cubes, and {\it probe} them to decide the best variable to split on. Probing is the process of computing quality metrics (e.g., propagation rate) by running a SAT solver on the sub-formulas thus obtained.
\end{definition}

\begin{definition}[\bf Cubing problem]
Given a Boolean formula\/ $F(x_1,x_2,\dotsc,x_n)$ in conjunctive normal form (CNF), the cubing solver outputs a set $C=\{C_1,\dotsc,C_k\}$ of cubes defined over the variables $\{x_1,\dotsc,x_n\}$. The resulting sub-formulas $F \wedge C_i$ are defined as the \textit{partitions} or sub-formulas of the original formula $F$. 
\end{definition}

\begin{definition}[\bf Splitting or cubing tree]
A splitting (or cubing) tree is a full binary tree in which each node represents a subformula, and edges correspond to assignments of $\text{True}$ or $\text{False}$ to a selected splitting variable. A path from the root to a leaf defines a cube.
\end{definition}

\begin{definition}[\bf Cubing solver]
A cubing solver takes as an input a Boolean formula $F$ in conjunctive normal form (CNF) and explores the space of cubing trees to generate an effective set of cubes—ideally, those that minimize the total solving time when dispatched to conquering solvers.
\end{definition}

The primary value of lookahead heuristics lies in their ability to provide a \textit{global view} of the search space, in contrast to CDCL solvers that operate in a highly \textit{local} manner~\cite{nejati2020cdcl}. While lookahead incurs higher computational costs due to probing, it often leads to better partitions that dramatically reduce solving time. Empirical evidence has shown that CnC techniques can successfully solve hard SAT problems that remain unsolved by other approaches. For a comprehensive overview of CnC solvers and related heuristics, we refer readers to the Handbook of Satisfiability~\cite{biere2021look}.

\section{The Design and Implementation of \tool}

In this section, we present the design and implementation of {\tool}\@, our cubing solver based on MCTS. The challenge posed by the cubing problem involves navigating a vast search space of splitting trees of large depth, resembling the complexities found in strategic games like chess or Go. Exhaustive exploration of this extensive space is infeasible. Drawing inspiration from AlphaZero~\cite{silver2017mastering}, we address this challenge by truncating the search tree heuristically and approximating long-term outcomes using a \textit{value function}.

Unlike AlphaGo, which relies on neural networks to predict value and policy (requiring costly training), or traditional MCTS (relies on random rollouts and no prior success probabilities), \tool uses deductive reasoning via a SAT solver to compute symbolic rewards based on propagation rate. This enables \tool to efficiently prune the search space and make informed decisions at each step. While propagation rate is effective in our setting, the framework naturally extends to other solver-derived metrics. As the simulation progresses and the tree grows deeper, these symbolic signals guide \tool toward identifying promising variable splits that yield \textit{high reward} cubes.

\subsection{Input and Output}
The input to \tool is a CNF Formula $F$. The output of the cubing solver is a set of $k$ cubes which, respectively, in conjunction with $F$, produce sub-formulas $\{F_1, F_2, \dotsc, F_k\}$ to be given to $k$ worker solvers in parallel. The original instance is considered solved if at least one of the sub-formulas is determined to be satisfiable or all of them are determined to be unsatisfiable. 

\subsection{Problem Setup}
We formulate cubing as a deterministic Markov Decision Process (MDP), more specifically as a tree MDP~\cite{scavuzzo2022learning}. Each node corresponds to a CNF formula, and actions represent variable splits. Upon selecting a variable $x_i$, the environment transitions to two child states: $F \wedge x_i$ and $F \wedge \neg x_i$. The reward for each node is the propagation rate, which is a solver-derived metric.

The termination condition of a cube is a user-defined parameter $n_{\text{elim}}$, also known as the \textit{variable cutoff value}, which denotes that the splitting process at a particular node must stop if at least $n_{\text{elim}}$ variables have been eliminated (through propagation or as part of splitting variables) in the cube. This choice in defining the termination condition is motivated by the objective of attaining balanced cubes within the solving process~\cite{heule2011cube}. 

\subsection{Cubing Episode}
Each cubing episode begins at the root node, corresponding to the original CNF formula $F$, and proceeds to construct a binary splitting tree. Each node in this tree represents a subformula derived from a sequence of variable splits, and each edge corresponds to assigning a variable either true or false. At every node, \tool invokes an MCTS simulation to determine the most promising variable to split next, guided by solver-derived propagation scores. Once a variable $x_i$ is selected, the tree is expanded by creating two child nodes corresponding to the formulas $F \wedge x_i$ and $F \wedge \neg x_i$. This process is repeated recursively on each child node unless the termination condition is met. Once the tree is fully expanded, \tool extracts a set of final cubes from the root-to-leaf paths. Each cube corresponds to a conjunction of literals encountered along a path, and collectively, these cubes define a partition of the original formula into sub-formulas suitable for parallel solving.

\subsection{MCTS Simulation}
The cubing episode invokes the MCTS simulation, generating an MCTS tree with the root node representing the formula passed by the cubing episode. Unlike standard MCTS approaches that rely on random rollouts or learned policy and value functions, \tool performs symbolic selection and rollouts guided entirely by deductive reasoning. Specifically, it uses SAT solvers with BCP to evaluate each node and compute propagation rates. This allows \tool to obtain accurate, domain-specific feedback without relying on learned approximations. A user-defined simulation budget controls the number of MCTS simulations, each comprising four phases: \emph{Selection}, \emph{Expansion}, \emph{Rollout}, and \emph{Backup}. The simulation budget was chosen to be 10 using light tuning on a smaller KS instance and then fixed for all benchmarks to balance cubing cost and exploration depth.

\subsubsection{Selection}

We employ the PUCT algorithm~\cite{rosin2011multi,silver2017mastering} to guide the tree traversal during the selection phase. Rather than using a neural network to provide priors over actions, \tool uses deductively computed scores based on BCP. For each valid variable $x_i$ (i.e., one that has not yet been split or propagated), we compute the propagation rate in both the positive and negative directions using BCP. These are combined into a scalar value using the formula used by \march~\cite{heule2011cube}: $\prop(x_i) \cdot \prop(\neg x_i) + \prop(x_i) + \prop(\neg x_i)$.

We normalize the score, and it becomes the prior probability $P(s,a)$ of taking action $a$ (splitting on variable $x_i$) in state $s$. Action selection is then guided by:
\begin{align}
a_{\text{chosen}} &= \arg \max_a \left(Q(s, a) + u(s, a)\right) \\
u(s, a) &= c_{\text{puct}} \cdot P(s, a) \cdot
\frac{\sqrt{\sum_b N(s, b)}}{1 + N(s, a)}
\end{align}
Here, $Q(s,a)$ is the expected reward of action $a$, $N(s,a)$ is the visit count, and $c_{\text{puct}}$ controls the exploration-exploitation trade-off. 

\subsubsection{Expansion} 
If the current node $s$ is non-terminal and the selected action $a$ is unvisited, then we create children nodes $s \wedge a$ and $s \wedge \neg a$. This allows the search tree to systematically explore promising branches guided by symbolic reasoning.

\subsubsection{Rollout} 
Instead of relying on random rollouts (which are expensive in SAT due to repeated BCP calls), \tool terminates simulations early and leverages deductive rewards available at intermediate nodes. That is, even non-terminal nodes can be evaluated based on how many variables have been eliminated via propagation. This intermediate feedback allows us to avoid full traversal to termination while still receiving informative reward signals.

\subsubsection{Backup}
During the backup phase, we propagate the reward values from children to their parents. We compute the value at each non-leaf node as the average of its children, i.e., values from the true and false branches. This approach allows the MCTS tree to reason about the cumulative impact of variable choices over deeper levels of the search.

\subsubsection{Final Action Choice}

Finally, we choose the action that has achieved the highest reward in any simulation: $a^* = \arg \max_a B(s,a)$. Here, $B(s,a)$ tracks the maximum (long-term) reward observed for action $a$ at state $s$. This design is well suited for our case, where the goal is to find a single variable split that leads to the most effective cube.

\section{Experimental Setup}

\subsection{Competing Tool} \label{march}

For our comparative analysis, we benchmark \tool against the well-established CnC cubing solver \march~\cite{heule2011cube}, which remains state-of-the-art in this domain. \March selects splitting variables using heuristics based on propagation count, certain characteristics of the formula, and related deductive metrics. Remarkably, for at least 15 years \march has been the most widely adopted CnC cubing solver, and has been employed in a range of combinatorial settings \cite{heule2018schur,bright2022satisfiability,semenov2023using,likochenspecker}. As described in Section~\ref{bg_cnc}, the \march cubing strategy exemplifies the practical success of CnC methods on hard symbolic problems.

Furthermore, recent SAT Competitions (2022 and 2023) have featured solvers such as Paracooba~\cite{heisinger2022paracooba} and Parallel MergeSAT~\cite{manthey2023parallel}, both of which integrate \march as their lookahead cubing component. This continued reliance underscores {\march}’s dominance and relevance as the cutting-edge CnC solver in the SAT ecosystem.

\subsection{Benchmarks} \label{benchmarks}

We evaluate the performance of \march and \tool on a diverse range of combinatorial benchmarks. Due to differing edge variable ordering conventions in their symmetry propagators, we cannot use SAT+CAS encodings with SMS, or vice versa. Consequently, some benchmarks are used to evaluate only one SAT+CAS or SMS, depending on the availability of the encoding. For the \march baseline, we followed configurations reported in prior CnC studies that used \march on these benchmarks. We used the same variable cutoff ($n_{\text{elim}}$) range as prior CnC studies.

\subsubsection{The Minimum Kochen--Specker Problem}

The minimum Kochen--Specker (KS) problem is of significant importance in quantum mechanics and has captivated the attention of physicists and mathematicians for decades. At its core, the KS problem is intricately tied to the fundamental principles of quantum mechanics, specifically addressing the notion of contextuality.

For using SAT+CAS~\cite{zulkoski2015mathcheck,bright2022satisfiability} as the conquering solver, we use the SAT encodings and instances created by previous work in this domain~\cite{scsquare} that used Cube-and-Conquer (using \march as their cubing solver). We replace \march with our \tool solver for the cubing process for an apples-to-apples comparison. We focus on CNF instances corresponding to KS order 19 (with 3,876 variables and 233,219 clauses), order 20 (with 4,560 variables and 408,455 clauses), order 21 (with 5,320 variables and 923,933 clauses), and order 22 (with 6,160 variables and 2,496,012 clauses) because of their computation tractability. For additional background on the KS problem and its SAT encoding, we refer our readers to the previous work~\cite{likochenspecker}.

For using SMS~\cite{kirchweger2023co} as a conquering solver, we generated the instances following its documentation. A key distinction in our approach is that the graph coloring property is specified dynamically via an external propagator rather than being statically encoded. Consequently, the resulting CNF file is substantially smaller. Specifically, order~20 contains 2,550 variables and 23,915 clauses, order~21 contains 2,884 variables and 28,609 clauses, and order~22 contains 3,245 variables and 33,979 clauses. For further details on the encoding, we refer the reader to the original paper~\cite{kirchweger2023co}.

\subsubsection{The Ramsey Problem}

The foundation of Ramsey Theory was laid by Frank P. Ramsey~\cite{ramsey1987problem}, with Ramsey numbers standing as renowned and challenging problems. Only nine non-trivial Ramsey numbers are currently known, despite extensive computational efforts~\cite{radziszowski2011small}. The Ramsey problem seeks to determine the smallest integer $n$ such that every red/blue coloring of the complete graph on $n$ vertices contains either a blue triangle or a red $k$-clique. 

In this benchmark, we specifically focus on the Ramsey problem of $R(8,3)=28$. To formulate these problems as SAT instances, we use the encodings from prior work~\cite{liramsey}, which include symmetry-breaking clauses and constraints informed by theoretical insights in Ramsey theory. The resulting CNF instances contain 15,820 variables and 3,163,013 clauses for $R(8,3)$.

\subsubsection{The Murty--Simon Conjecture}

The diameter of a graph \(G\), denoted \(\text{diam}(G)\), is the maximum distance between any pair of vertices, where the distance between two vertices is defined as the length of the shortest path connecting them. A graph is called \emph{diameter-\(d\)-critical} if it has diameter \(d\) and the removal of any edge strictly increases its diameter. 

The investigation of diameter-\(d\)-critical graphs originated with the seminal work of Erdős and Rényi~\cite{erdHos1966problem}. Of particular importance is the case when \(d = 2\), which is connected to the Murty--Simon Conjecture. This conjecture states that any diameter-2-critical graph with \(n\) vertices and \(m\) edges satisfies \(m \le \lfloor n^2 / 4 \rfloor\), with equality holding if and only if the graph is a complete bipartite graph \(K_{\lceil n/2 \rceil, \lfloor n/2 \rfloor}\). 

We employ the SAT encoding introduced in prior work by Kirchweger and Szeider~\cite{kirchweger2021sat}. Specifically, we use a CNF formula \(D_2(n, m)\) that is satisfiable if and only if there exists a diameter-2-critical graph \(G\) with \(n\) vertices and \(m\) edges. We choose the instance $D_2(15,56)$, which contains 11,655 variables and 71,338 clauses.

\subsection{Implementation and Computational Environment}

Our implementation is written in Python 3.11. We leverage MiniSAT's \texttt{propagate} functionality, integrated via the PySAT library~\cite{pysat}, to compute the propagation rate through unit propagation. All experiments were conducted on a high-performance cluster equipped with dual AMD EPYC 7713 CPUs running at 2.0~GHz, providing a total of 128 cores per node and 512GB DDR4 DRAM\@. The cluster nodes run on Red Hat Enterprise Linux 9.4 (Plow).

\subsection{Cubing and Solving using \MC} \label{mathcheck}

Our experimental setup makes use of the \MC pipeline~\cite{likochenspecker}, which manages the complete Cube-and-Conquer workflow, including cubing, simplification, parallel solving, and dynamic task allocation for efficiently tackling large combinatorial CNF instances. A key feature of \MC is its \textit{iterative} approach to cubing and solving: rather than generating all cubes upfront, it incrementally produces cubes and dispatches them immediately to parallel worker solvers. This design has two major benefits: (1) It enables the cubing process to be parallelized, unlike traditional cubing solvers that operate sequentially. (2) It allows \textsc{CaDiCaL}~\cite{Biere2024} to simplify the formula after each cubing decision, so subsequent cubing steps operate on a progressively simplified formula.

The pipeline also includes automated timeout handling. If a worker solver exceeds a specified time limit when attempting to solve a subproblem, that subproblem can be sent back for further cubing, effectively subdividing it into easier instances. This adaptive mechanism ensures that harder cubes do not become bottlenecks in the solving process and enhances load balancing.

\MC supports seamless switching between different cubing and conquering solvers. In our experiments, we switch between the default \march cubing solver and our method, \tool, while keeping the rest of the pipeline unchanged. The resulting cubes are solved using two widely adopted conquering solvers for combinatorial problems, \emph{SAT+CAS} and \emph{SMS}.
\begin{enumerate}
    \item \textbf{SAT + Computer Algebra Systems (SAT+CAS)}~\cite{satcas}, which has been used for combinatorial problems such as Lam's problem~\cite{bright2021sat}, the Kochen--Specker problem~\cite{likochenspecker}, the Ramsey problem~\cite{liramsey}, etc. In each use case, the CASs provide a variety of combinatorial knowledge to the SAT solver in clausal form.
    \item \textbf{SAT Modulo Symmetries (SMS)}~\cite{satmsym}, which augments a SAT solver with a custom propagator capable of reasoning about symmetries. This framework has been used to tackle problems like the Kochen--Specker problem~\cite{kirchweger2023co}, the Erdős–Faber–Lovász conjecture~\cite{kirchwegererdos}, and Rota’s Basis Conjecture~\cite{kirchwegerrota}. We integrated SMS into MathCheck as a conquering solver. While SMS provides a cubing pipeline by calling \march~\cite{kirchweger2025smart} and then solving the cubes in parallel, we use the \MC cubing pipeline instead so that the cubing process can be parallelized.
\end{enumerate}
\MC coordinates iterative cubing, parallel solving, timeout handling, and dynamic load balancing to ensure high CPU utilization. By integrating \tool into \MC, we demonstrate its effectiveness as a drop-in replacement that enhances the cubing phase while remaining compatible with a wide range of downstream conquering solvers. This leads to improved overall Cube-and-Conquer performance on challenging combinatorial benchmarks.

\begin{table*}[ht]
\centering
\small
\setlength{\tabcolsep}{3pt}
\resizebox{0.7\textwidth}{!}{
\begin{tabular}{ccccccc}
\hline
\textbf{\begin{tabular}[c]{@{}c@{}}Instances\end{tabular}} &
\textbf{\begin{tabular}[c]{@{}c@{}}Tools\end{tabular}} &
\textbf{\begin{tabular}[c]{@{}c@{}}Cubing and \\ simplification \\ CPU time (h)\end{tabular}} &
\textbf{\begin{tabular}[c]{@{}c@{}}Total \\ CPU time (h)\end{tabular}} &
\textbf{\begin{tabular}[c]{@{}c@{}}Total \\ CPU time \\ speedup\end{tabular}} &
\textbf{\begin{tabular}[c]{@{}c@{}}Total \\ elapsed \\ real time (h)\end{tabular}} &
\textbf{\begin{tabular}[c]{@{}c@{}}Total elapsed \\ real time \\ speedup\end{tabular}} \\ \hline
  
\multirow{2}{*}{Ramsey (8,3)} 
  & \march &  3.49  &  9.36  &    &  2.16  &    \\
  & \textsc{AMS} &  \textbf{1.22}  &  \textbf{7.58}  &  \textbf{1.23$\times$}  &  \textbf{0.94}  &  \textbf{2.30$\times$}  \\ \hline
  
\multirow{2}{*}{KS 19}
  & \march &  1.16  &  1.70  &    &  0.17  &    \\
  & \textsc{AMS} &  \textbf{0.29}  &  \textbf{0.54}  &  \textbf{3.15$\times$}  &  \textbf{0.08}  &  \textbf{2.13$\times$}  \\ \hline

\multirow{2}{*}{KS 20}         
  & \march &  3.59  &  7.61  &   &  0.60  &   \\
  & \textsc{AMS} &  \textbf{0.90}  &  \textbf{3.21}  &  \textbf{2.37$\times$}  &  \textbf{0.31}  &  \textbf{1.93$\times$}  \\ \hline
  
\multirow{2}{*}{KS 21}         
  & \march &  104.88  &  253.94  &    &  5.59  &    \\
  & \textsc{AMS} &  \textbf{5.64}  &  \textbf{49.48}  &  \textbf{5.13$\times$}  &  \textbf{1.74}  &  \textbf{3.21$\times$}  \\ \hline
  
\multirow{2}{*}{KS 22}         
  & \march &  & timeout & & timeout & \\
  & \textsc{AMS} &  \textbf{243.67}  &  \textbf{1841.18}  &    &  \textbf{16.34}  &    \\ \hline
  
\end{tabular}
}
\caption{\textbf{Parallel Cubing and Parallel Solving (SAT+CAS):} Comparison results between \march and \tool (\textsc{AMS}) across various benchmarks. Parallel cubing is performed using \march and \tool, while parallel solving is performed using SAT+CAS on a 128 core machine. The timeout was set to 5 days. Note that the Murty--Simon Conjecture is excluded due to encoding incompatibilities with the SAT+CAS symmetry propagator.} 

\label{tab:main}
\end{table*}

\subsection{Metrics} \label{metrics}

The evaluation metrics used in our study follow established practices in the CnC literature~\cite{heule2011cube,likochenspecker}, providing a robust basis for comparing solver performance. We focus on the following two key metrics:

\begin{itemize}[leftmargin=1.25em]
    \item \textbf{Total CPU time:} The aggregate time spent across all CPU cores for cubing, simplification, and solving. This metric reflects the overall computational cost and resource utilization of the (parallel) CnC solvers under evaluation.
    
    \item \textbf{Elapsed real time:} The end-to-end wall clock time taken to complete cubing, simplification, and solving for a given benchmark. This metric captures the latency or time-to-solution experienced by the user. (Note: elapsed real time excludes cluster job scheduling overheads.)
\end{itemize}

\section{Experimental Results}

\vspace{0.1cm}

In this section, we present a detailed empirical evaluation of \tool against the state-of-the-art \march cubing solver (Section~\ref{march}) on challenging combinatorial instances described in Section~\ref{benchmarks}.  We evaluate \tool in conjunction with two distinct conquering solvers: (1) SAT+CAS and (2) SMS. Our results demonstrate that \tool consistently outperforms \march, providing significant improvements in elapsed real time across all benchmarks, irrespective of the chosen conquering solver. We also analyze cube-level behavior and parallel scaling, and present them in the Sections~\ref{cubeanalyze} and~\ref{scalingstudies}.

For each instance, we replaced {\march}’s cubing solver with \tool with the same command-line arguments (e.g., variable cutoff conditions) while keeping the downstream solving pipeline unchanged. Specifically, we used the parallel cubing and solving framework provided by \MC (Section~\ref{mathcheck}), which allows for an apples-to-apples comparison under identical conditions.

\subsection{Results with SAT+CAS}

Table~\ref{tab:main} reports the results when SAT+CAS is used as the conquering solver. \tool consistently outperforms \march across all metrics. The performance gap becomes increasingly pronounced on harder instances. For example, on the KS order 21 instance, where \tool takes substantially less amount of time in cubing as compared to \march (18$\times$ speedup on KS order 21), and still achieves a 5.13$\times$ speedup in total CPU time and a 3.21$\times$ speedup in elapsed real time compared to \march. On the most challenging KS order 22 instance, the conquering solver times out after five days when using cubes generated by \march, whereas the cubes generated by \tool are successfully solved in under 17 hours of elapsed real time.

The observed performance improvements are consistent with \tool’s use of MCTS to guide cubing. Both \tool and \march use the same preprocessing and variable preselection, with the primary algorithmic difference being the MCTS-based exploration strategy. While \march selects splitting variables using greedy heuristics~\cite{heule2011cube}, it limits search depth to reduce overhead, often resulting in sub-optimal partitions. In contrast, {\tool}'s informed exploration leads to better cube quality, reducing both CPU time and wall clock time. These gains arise from improved cube quality, as \tool generates fewer cubes than \march for the same variable cutoff and avoids a small number of disproportionately hard cubes that dominate elapsed time (Section~\ref{cubeanalyze}).

\subsection{Results with SMS}

\begin{table}[ht]
\centering
\small
\setlength{\tabcolsep}{1pt}
\begin{tabular}{cccccc}
\hline
\textbf{\begin{tabular}[c]{@{}c@{}}Instances\end{tabular}} &
\textbf{\begin{tabular}[c]{@{}c@{}}Tools\end{tabular}} &
\textbf{\begin{tabular}[c]{@{}c@{}}Cubing \\ CPU \\ time (h)\end{tabular}} &
\textbf{\begin{tabular}[c]{@{}c@{}}Total \\ CPU \\ time (h)\end{tabular}} &
\textbf{\begin{tabular}[c]{@{}c@{}}Total \\ elapsed \\ real \\ time (h)\end{tabular}} &
\textbf{\begin{tabular}[c]{@{}c@{}}Total elapsed \\ real time \\ speedup\end{tabular}} \\ \hline

\multirow{2}{*}{$D_2(15,56)$} 
  & \march &  \textbf{0.07}  &  \textbf{2.81}  &  0.37  &   \\
  & \textsc{AMS} &  1.98  &  3.87  &  \textbf{0.23}  &  \textbf{1.61$\times$}  \\ \hline

\multirow{2}{*}{KS 20} 
  & \march &  \textbf{0.01}  &  \textbf{0.51}  &  0.50  &   \\
  & \textsc{AMS} &  0.57  &  1.41  &  \textbf{0.16}  &  \textbf{3.12$\times$}  \\ \hline

\multirow{2}{*}{KS 21} 
  & \march &  \textbf{0.04}  &  8.15  &  8.18  &   \\
  & \textsc{AMS} &  0.65  &  \textbf{7.70}  &  \textbf{1.08}  &  \textbf{7.57$\times$}  \\ \hline

\multirow{2}{*}{KS 22} 
  & \march & & timeout & timeout &  \\
  & \textsc{AMS} &  \textbf{18.08}  &  \textbf{137.39}  &  \textbf{10.01}  &    \\ \hline

\end{tabular}
\caption{\textbf{Parallel Cubing and Parallel Solving (SMS):} Comparison results between \march and \tool (\textsc{AMS}) across various benchmarks. Parallel cubing is performed using \march and \tool, while parallel solving is performed using SMS conquering solver on a 128 core machine. The timeout was set to 5 days. Note that Ramsey (8,3) is excluded due to encoding incompatibilities with the SMS symmetry propagator.}
\label{tab:sms}
\end{table}

Table~\ref{tab:sms} reports the results when SMS is used as the conquering solver. Similar to the SAT+CAS setting, \tool consistently outperforms \march on elapsed real time. On KS order 21, \tool achieves an \smsspeedupx speedup in elapsed real time. For KS order 22, the conquering solver times out after five days when using cubes generated by \march, whereas the cubes generated by \tool are successfully solved in around 10 hours. As in the SAT+CAS case, these gains are explained by cube-quality effects. \March produces unbalanced cubes that limit parallel scalability on harder instances despite lower CPU usage on easier instances, while \tool avoids disproportionately hard cubes and, as instance difficulty increases, outperforms \march in both total CPU time and elapsed real time.

\subsection{Cube-Level Analysis on KS Order 21}~\label{cubeanalyze}

\begin{figure}[h]
  \centering
  \includegraphics[width=\linewidth]{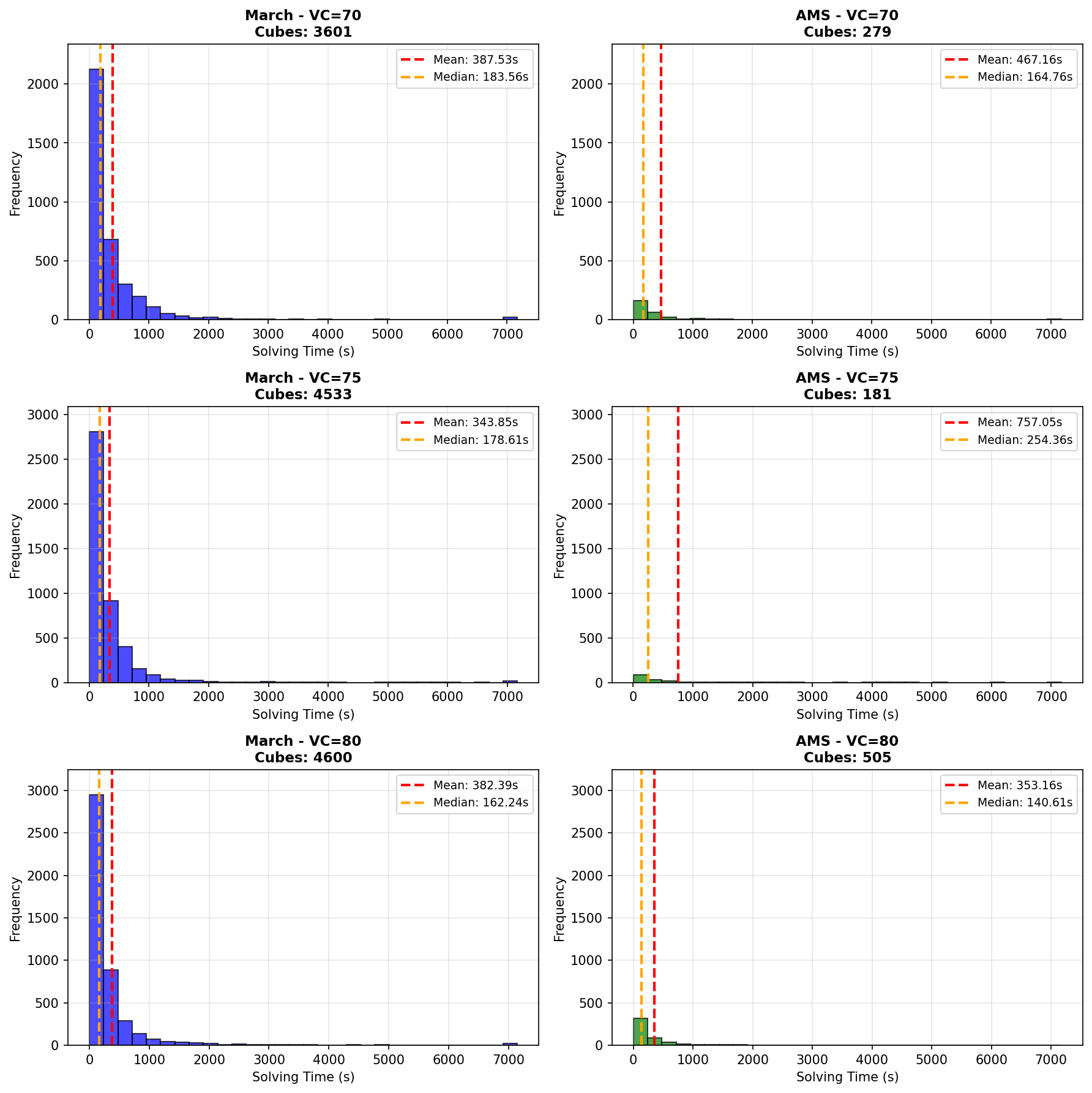}
  \caption{Distribution of per-cube CPU solving times for the KS order 21 instance with SAT+CAS on 128 cores, comparing \march and \tool across different variable cutoffs (VC). The figure also reports the number of cubes, along with mean and median cube solving times.}
  \label{fig:cube-dist-ks21}
\end{figure}

We analyze cube-level behavior of \tool against \march on the KS order 21 instance with SAT+CAS to better explain why \tool outperforms \march. By cube-level analysis we mean that we look at the solving time corresponding to different cubes generated by both the tools. Put differently, we generate sets of cubes using each tool, and then compare the average solving time over these cubes, while keeping the input problem fixed. We chose the KS 21 instance because it was the hardest benchmark, in terms of total elapsed real time and total CPU time, that was solved successfully by both \march and \tool.

Figure~\ref{fig:cube-dist-ks21} shows the distribution of per-cube solving times on a 128-core machine for different variable cutoffs. Across all cutoffs, \tool produces substantially fewer cubes than \march. More importantly, \tool avoids generating a large number of extremely time-consuming cubes. In contrast, \march consistently produces many trivial cubes along with a non-negligible number of very hard cubes that dominate the total solving time.

\begin{figure}[t]
  \centering
  \includegraphics[width=\linewidth]{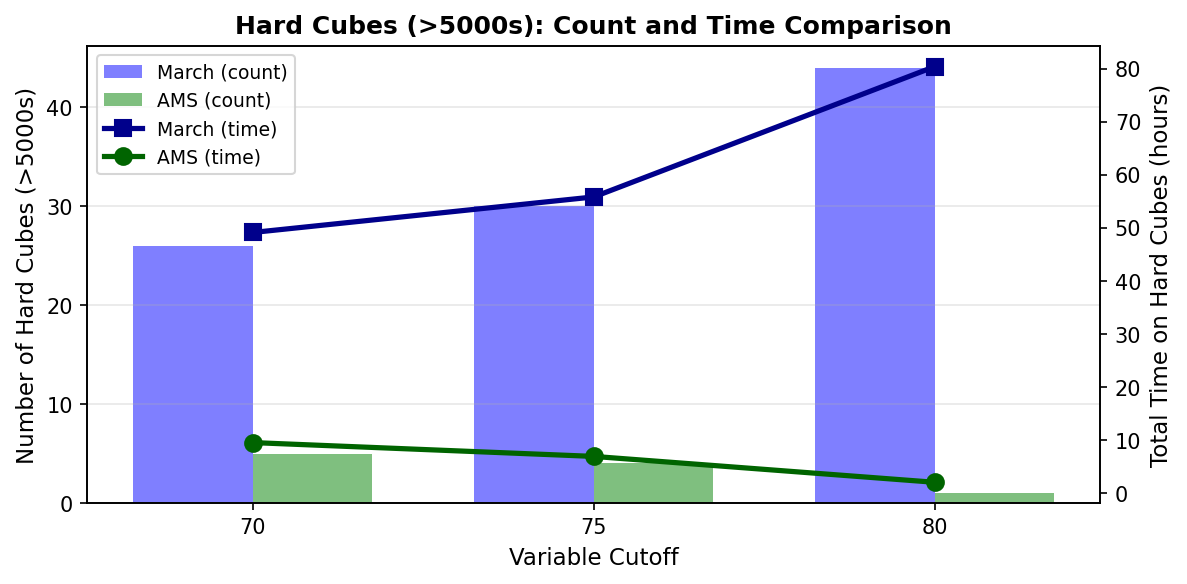}
  \caption{Number of hard cubes (solving time $>5000$s) and cumulative time spent on these cubes for the KS order 21 instance with SAT+CAS on 128 cores. \tool generates significantly fewer hard cubes and spends substantially less time on them than \march.}
  \label{fig:hard-cubes-ks21}
\end{figure}

This effect is made explicit in Figure~\ref{fig:hard-cubes-ks21}, which focuses on cubes whose solving time exceeds 5000 seconds. \tool does not merely reduce the number of cubes uniformly, but specifically avoids generating the hardest cubes that consume the majority of solving time. \march, on the other hand, generates significantly more such hard cubes, and the cumulative time spent on them increases with the variable cutoff. We argue that this behavior explains why \tool achieves superior elapsed real time in Tables~\ref{tab:main} and~\ref{tab:sms}, even in cases where total CPU time is comparable or slightly higher.

\subsection{Parallel Scaling Behavior}~\label{scalingstudies}

\begin{figure}[h]
  \centering
  \includegraphics[width=\linewidth]{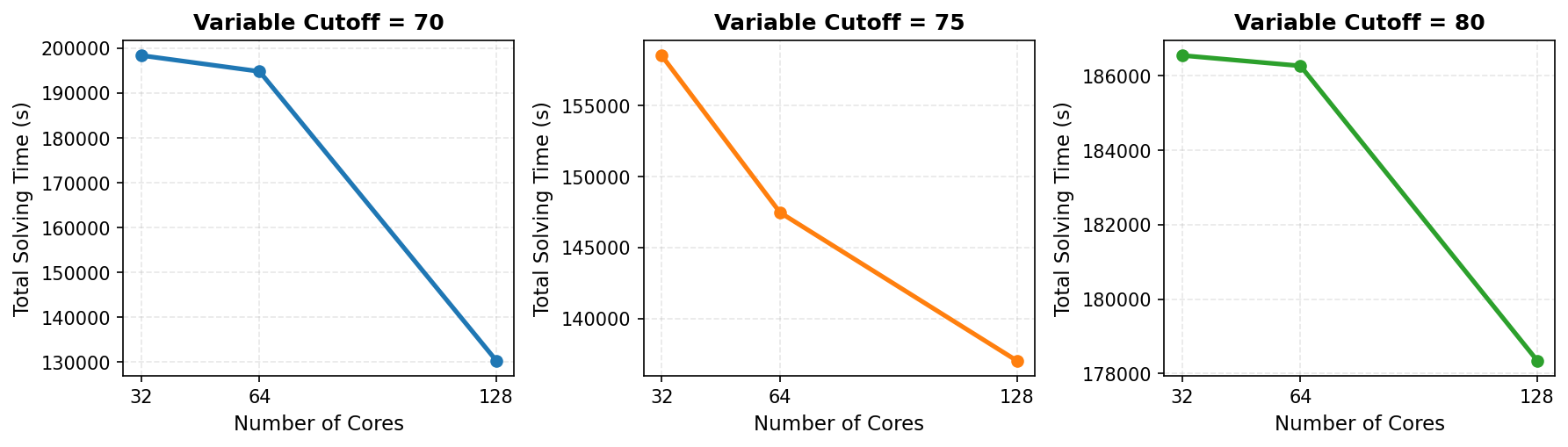}
  \caption{Parallel scaling behavior on the KS order 21 instance with SAT+CAS. Total CPU time is shown as a function of the number of cores for different variable cutoffs. \tool scales consistently with increasing core counts.}
  \label{fig:scaling-ks21}
\end{figure}

We further evaluate parallel scaling on the KS order 21 instance with SAT+CAS by varying the number of cores (32, 64, and 128 cores) while fixing the variable cutoff. Figure~\ref{fig:scaling-ks21} reports total CPU time as a function of the number of cores. \tool exhibits good scaling behavior across all variable cutoffs, with total solving time decreasing as additional cores are made available. This indicates that the cubes generated by \tool are well suited for parallel execution.

\subsection{Summary}

These results demonstrate that {\tool}'s benefits are agnostic to the choice of the conquering solver and translate to both the SAT+CAS and SMS pipelines. The performance improvements can be attributed to {\tool}'s use of MCTS to guide cubing. While \march relies on greedy heuristics with limited search depth~\cite{heule2011cube}, often yielding sub-optimal partitions, {\tool}'s informed exploration produces higher-quality cubes. As a result, both total CPU time and wall clock time are reduced, and we further show that these gains arise from improved cube distributions and more robust parallel scaling.

\section{Conclusion, Limitation, and Future Work}

In this paper we present \tool, a novel deductive feedback-based MCTS cubing solver. Using an informed exploration of the search space, our tool overcomes the limitations of both greedy (limited search resulting in sub-optimal cubes) and exhaustive approaches (search cost can overwhelm any benefits derived from the computation of optimal cubes). MCTS guided by deductive reward improves the search process for better decision-making across diverse combinatorial problems. Our approach effectively balances exploration and exploitation to generate high-quality cubes while keeping cubing overhead low.

Our experimental results show substantial performance improvements over the state-of-the-art \march cubing solver. We show that in all cases, there is a speedup in elapsed real time ranging from \minspeedupx to \maxspeedupx, driven by improved cube-level behavior and scaling properties, on a 128 core machine across three challenging combinatorial benchmarks: the minimum Kochen--Specker problem, the Murty--Simon Conjecture, and the Ramsey problem.

While \tool shows substantial performance improvements over existing cubing solvers such as \march, there remains room for further optimization. The current Python implementation relies on PySAT to invoke the underlying SAT solver for deductive rewards via BCP, which introduces additional overhead compared to C-based solvers like \march. Moreover, \tool may face scalability challenges on instances with very large number of variables to split on, where the increased branching factor can reduce the efficiency of MCTS exploration.

Future work includes developing a C++ implementation of \tool to reduce runtime overhead and improve integration with SAT solvers. We also plan to explore deep learning–based MCTS techniques for cubing, with the goal of adapting across instances from the same problem class and learning more efficient cubing strategies.

\bibliographystyle{named}
\bibliography{main}

\end{document}